\documentclass[letterpaper,10pt,conference]{ieeeconf}
\usepackage{etoolbox}
\usepackage{graphicx}
\usepackage{arydshln}
\usepackage{amsmath}
\usepackage{amssymb}
\usepackage{enumerate}
\usepackage{booktabs}
\usepackage{authblk}
\usepackage{color}
\usepackage{url}
\usepackage{multirow}
\usepackage[table]{xcolor}
\usepackage{stfloats}
\usepackage{multirow}
\usepackage{booktabs}
\usepackage{microtype}
\usepackage[pagebackref=true,breaklinks=true,letterpaper=true,colorlinks,bookmarks=false]{hyperref}
\let\labelindent\relax
\usepackage{enumitem}
\setlist[itemize]{leftmargin=*}

\usepackage[font=footnotesize]{caption}

\IEEEoverridecommandlockouts
\begin{document}
\title{MarineDet: Towards Open-Marine Object Detection}
\author{Haixin~Liang$^{1\dagger}$,~Ziqiang~Zheng$^{2\dagger*}$,~Zeyu~Ma$^{3}$,~Sai-Kit~Yeung$^{1,2}$
 \thanks{$^1$Haixin Liang and Sai-Kit Yeung are with the Division of Integrative Systems and Design, Hong Kong University of Science and Technology. 
 }
 \thanks{$^2$Ziqiang Zheng and Sai-Kit Yeung are with the Department of Computer Science and Engineering, The Hong Kong University of Science and Technology.}
 \thanks{$^3$Zeyu Ma is with the School of Computer Science and Engineering, University of Electronic Science and Technology of China}
 \thanks{$^{\dagger}$These two authors contributed to this work equally.}
 \thanks{$^*$Corresponding author: Ziqiang Zheng (zhengziqiang1@gmail.com)}
 \vspace{-2em}
}
\vspace{-0.4in}
\let\oldtwocolumn\twocolumn
\renewcommand\twocolumn[1][]{%
    \oldtwocolumn[{#1}{
    \begin{center}
           \vspace{-1.5em}
           \includegraphics[width=\textwidth]{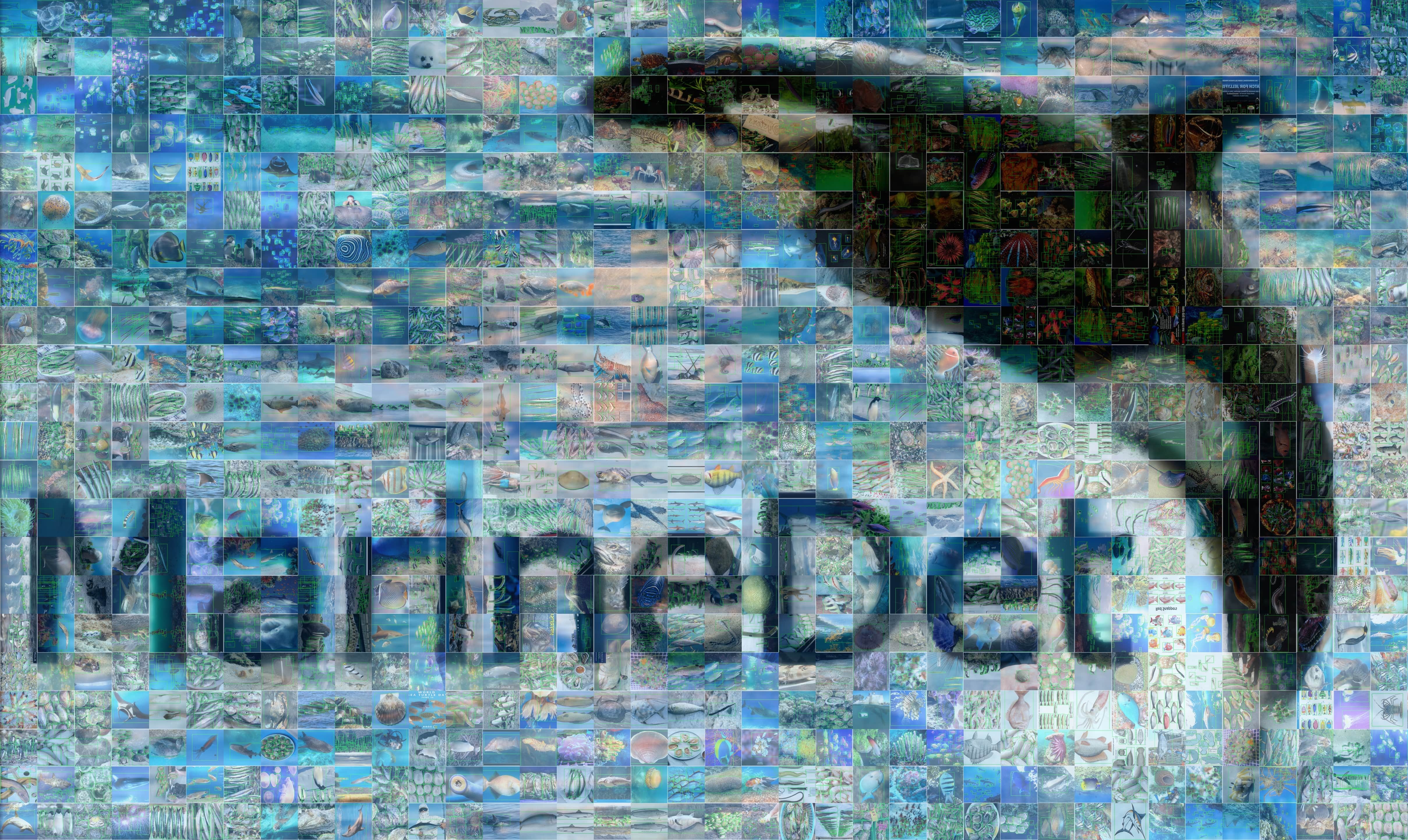}
           \captionof{figure}{Example images with bounding box annotations from our MarineDet dataset.}
           \label{fig:teaser}
           \vspace{-0.5em}
        \end{center}
    }]
}

\maketitle
\makeatletter
\patchcmd{\@makecaption}
  {\scshape}
  {}
  {}
  {}
\makeatother
\begin{abstract}
Marine object detection has gained prominence in marine research, driven by the pressing need to unravel oceanic mysteries and enhance our understanding of invaluable marine ecosystems. There is a profound requirement to efficiently and accurately identify and localize \textit{diverse} and \textit{unseen} marine entities within underwater imagery. The open-marine object detection (\textbf{OMOD} for short) is required to detect diverse and unseen marine objects, performing categorization and localization simultaneously. To achieve OMOD, we present \textbf{MarineDet}. We formulate a joint \textit{visual-text semantic space} through pre-training and then perform \textit{marine-specific training} to achieve in-air-to-marine knowledge transfer. Considering there is no specific dataset designed for OMOD, we construct a \textbf{MarineDet dataset} consisting of 821 marine-relative object categories to promote and measure OMOD performance. The experimental results demonstrate the superior performance of MarineDet over existing generalist and specialist object detection algorithms. To the best of our knowledge, we are the first to present OMOD, which holds a more valuable and practical setting for marine ecosystem monitoring and management. Our research not only pushes the boundaries of marine understanding but also offers a standard pipeline for OMOD.
\end{abstract}

\section{Introduction}
Marine ecosystems are vital components of the health of our planet, playing a critical role in regulating climate, supporting biodiversity, and providing essential resources and services to both marine and terrestrial life. Performing marine object detection~\cite{williams2011adaptive,fulton2019robotic,zocco2023towards,xia2019visual,bovcon2019mastr1325,lyu2022sea} is essential for comprehending the intricate dynamics of underwater ecosystems, enabling accurate species identification, behavior analysis, and ecosystem health assessment. Marine object detection could also support thorough biodiversity assessments~\cite{kuenzer2014earth}, tracking and monitoring the movements of marine species, comprehending ecosystem dynamics, facilitating data-driven insights into the health and behavior of underwater environments, guiding conservation strategies, and bolstering the ability to protect fragile marine habitats.

\begin{figure}[t]
  \centering
    \includegraphics[width=\linewidth]{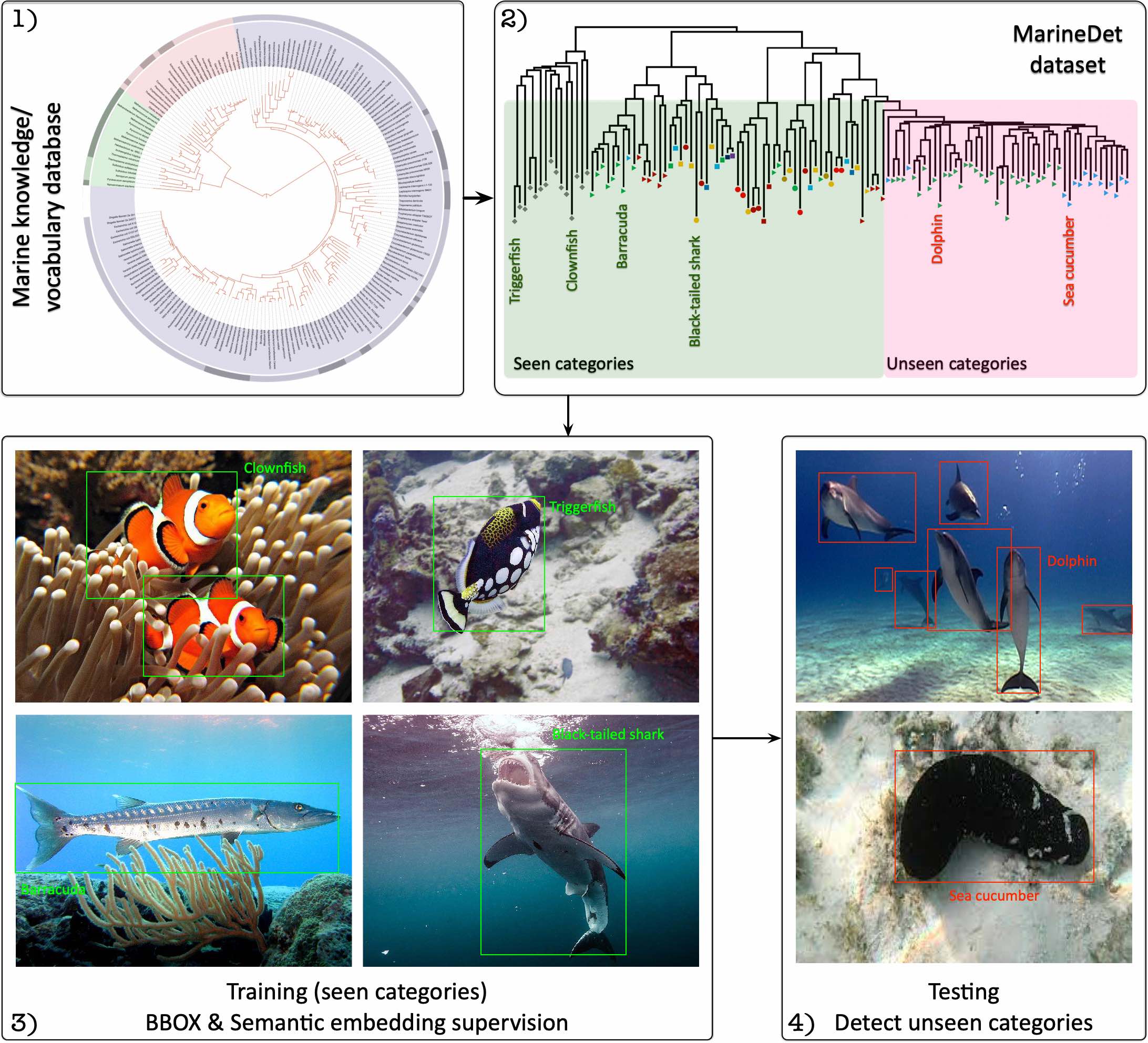}
  \caption{The demonstration of open-marine object detection. To achieve effective open-marine object detection, we first construct a large scale marine vocabulary databased to include a large range of marine object conceptions in 1), then we propose a large scale \textbf{MarineDet dataset} for performing open-vocabulary object detection in 2). During the training procedure, the bounding box annotations and semantic embedding supervision from the \textit{seen} categories are provided for optimizing the open-vocabulary object detection models in 3). At the final inference stage represented in 4), the trained model could detect the \textit{unseen} categories based on the hierarchical relationships presented in the constructed marine knowledge database.}
  \label{fig:demo}
\end{figure}

However, the existing marine object detection algorithms~\cite{fan2020dual,yan2022underwater,kapoor2023underwater} are mainly limited to close-set object detection. The trained models can only recognize very few fixed pre-defined object categories. It is heavily constrained by its specificity to pre-defined target classes, potentially overlooking or misclassifying novel or rare species, artifacts, or object categories that are not part of the established training dataset. This limitation restricts its adaptability to evolving marine environments and hinders its effectiveness in scenarios where the identification of unforeseen objects is paramount. To address these issues, \textit{open-marine object detection} (\textbf{OMOD} for short) as demonstrated in Fig.~\ref{fig:demo} presents a unique and advantageous way for underwater analysis, offering a range of benefits that are pivotal in advancing our understanding of marine ecosystems. Unlike existing detection algorithms that focus on specific targets, OMOD allows for the identification of a wide spectrum of marine entities, including both \textit{seen} and \textit{unseen} marine object categories. Furthermore, OMOD contributes to biodiversity research by enabling the discovery of new and rare species, facilitating a more comprehensive assessment of ecosystem health and diversity.

Even though OMOD has overwhelming advantages, there are still some significant challenges. Performing OMOD poses challenges that span from identifying \textit{unseen} or \textit{rare} species, to adapting to dynamic and diverse underwater environments with varying lighting and underwater conditions. In this paper, we take the first attempt to perform OMOD and present \textbf{MarineDet}, a flexible, applicable, and versatile marine object detector for effectively detecting marine object categories. We formulate a joint \textbf{visual-text semantic space} through contrastive pre-training based on large-scale general-purpose image-text pairs. This pre-training facilitates the alignment of the regional visual features within the images and corresponding textual descriptions. After pre-training, the model has a strong ability to recognize a wide range of objects and discriminate the foreground categories from the background. Then we perform an \textbf{in-air-to-marine} knowledge transfer by transferring the learned pre-trained model to the marine domain. We demonstrate that such \textbf{marine-specific training} could empower the detection model with a strong ability to recognize both \textbf{seen} and \textbf{unseen} marine object categories. To promote and evaluate the performance of OMOD, we also present our \textbf{MarineDet dataset}, including 821 diverse marine-relative object categories with corresponding bounding box annotations. We have conducted comprehensive experiments based on different settings (\emph{e.g.}, fully-supervised, open-vocabulary, and fine-grained), exploring the effectiveness and boundaries of our MarineDet. The main contributions of this paper are as follows:
\begin{itemize}
    \item We are the first to perform open-marine object detection, empowering the detection model with the ability to detect a large range of seen and unseen marine objects.  
    \item We present our MarineDet dataset, the first dataset specially designed for OMOD. We scalably generate comprehensive attribute annotations based on ChatGPT-3.5/GPT-4. 
    \item Comprehensive experiments under various settings have been conducted to explore the effectiveness and boundaries of our MarineDet for OMOD. 
\end{itemize}

\section{Related Work}
\subsection{Marine Object Recognition}
Marine object recognition~\cite{fulton2019robotic,miller2019evaluating,sadrfaridpour2021detecting,lin2023oysternet} could unveil the mysteries of the oceans and harness technology to elevate marine research~\cite{li2021marine}, conservation~\cite{hong2020trashcan}, and industrial endeavors. Marine object detection provides a valuable tool for environmental monitoring and identifying shifts or disturbances in marine ecosystems. To enable marine object recognition, various datasets (\emph{e.g.}, MAS3K~\cite{li2020mas3k,li2021marine}, WildFish~\cite{zhuang2018wildfish}, WildFish++~\cite{zhuang2020wildfish++}, UDD~\cite{liu2021new}, SUIM~\cite{islam2020semantic}) have been proposed for promoting the recognition performance of marine organisms. However, most of the existing datasets only focus on a limited number of marine objects or only provide image-level annotations for fine-grained image classification. In this work, we aim to perform OMOD (recognizing and localizing objects meanwhile), which could not only detect a wide spectrum of seen marine object categories but also some unseen objects. 

\subsection{Open-vocabulary Object Detection}
Open-vocabulary detection (OVD)~\cite{zareian2021open,yao2023detclipv2,kim2023region,zohar2023prob,wang2023detecting} aims to generalize beyond the limited number of pre-fixed classes during the training phase. The goal is to detect novel classes defined by an unbounded (open) vocabulary at the inference stage. Unlike traditional closed-domain detection~\cite{ge2021yolox,NIPS2015,lu2019grid}, which focuses on limited predefined classes, OVD navigates the intricacies of recognizing both \textit{seen} and \textit{unseen} objects. This approach is pivotal for scenarios where novel objects or anomalies may arise, offering the potential to unlock new insights, enhance safety, and revolutionize fields such as autonomous driving~\cite{zheng2020forkgan}, surveillance~\cite{ferri2020cooperative}, environmental monitoring~\cite{karapetyan2018multi} and underwater exploring. The dominant way of performing OVD is to adopt a pre-trained visual encoder from a trained cross-modality alignment model (\emph{e.g.}, CLIP~\cite{radford2021learning} and BLIP~\cite{li2022blip,li2023blip}) which is optimized by millions of image-text pairs from public websites. OVR~\cite{zareian2021open} adopted the powerful pre-trained image encoder from CLIP and utilized it for detector initialization for better feature extraction on unseen object categories. RegionCLIP~\cite{kim2023region} proposed to perform the regional visual feature and the textual conception alignment to promote the generalization ability to unseen categories. However, there are relatively few or no research works that focus on open-marine object detection. In this work, we make the first attempt to perform OMOD.  

\begin{figure*}[t]
  \centering
  \includegraphics[width=\linewidth]{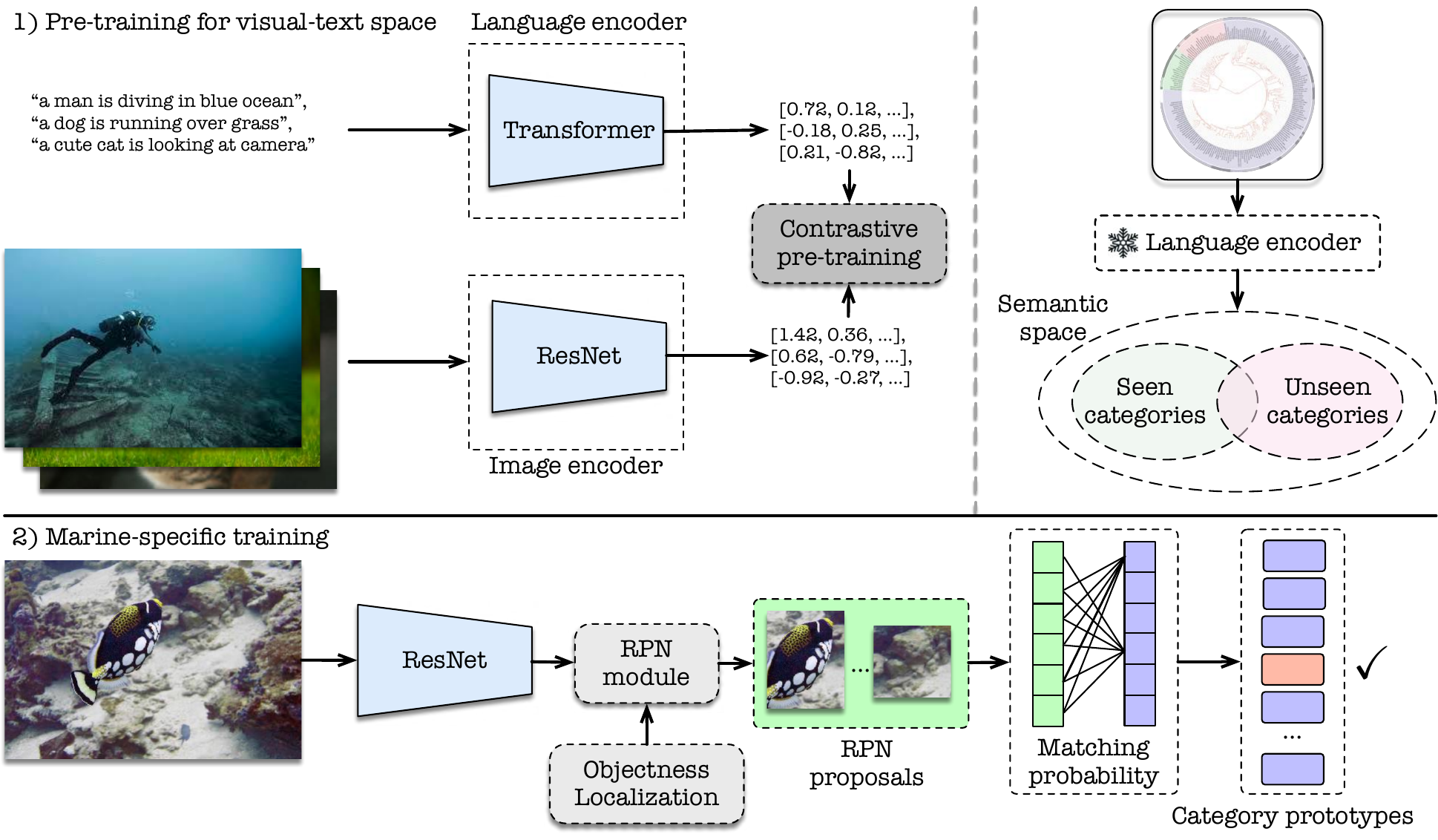}
  \caption{The framework overview of our MarineDet, where there are two main procedures: 1) \textbf{pre-training} for joint visual-text semantic space construction and 2) \textbf{marine-specific training}.}
  \label{fig:frame}
\end{figure*}

\section{Our Method}
\subsection{Overview}

\noindent\textbf{Problem Formulation}. We first provide preliminaries for open-marine object detection. We aim to develop algorithms that can accurately detect, localize, and categorize objects within images, even when faced with previously unseen object categories. Let $O$ represent the set of all possible object categories, where $|O| \rightarrow \infty$, indicating an open-ended vocabulary. We detect the presence of objects within the image by generating bounding boxes. We then assign category labels $C = \{c_i\}$ to the detected objects based on the similarity between the semantic textual embedding of $c_i$ and the regional features identified by the bounding box, where $c_i$ corresponds to the category label. To perform OMOD, we present \textbf{MarineDet}, as illustrated in Fig.~\ref{fig:frame}, where there are two main procedures: 1) \textbf{pre-training} for joint visual-text semantic space construction and 2) \textbf{marine-specific training}.

\subsection{Pre-training for Joint Visual-Text Semantic Space}
The multi-modal foundation model~\cite{radford2021learning,li2022blip,li2023blip} has demonstrated its powerful ability to construct a joint visual-text semantic space, where we could perform various downstream tasks. Through such constructed joint visual-text semantic space, we could include wide object conceptions and align the visual features and the corresponding textual descriptions. In this work, we take the first attempt to demonstrate the significance and valuability of introducing a constructed joint visual-text semantic space for effective open-marine object detection. During the pre-training procedure, we aim to compute the similarity between the extracted visual features from the RPN module and the generated textual conception based on contrastive learning~\cite{radford2021learning}:
\begin{equation}
\begin{array}{l}
\mathcal{L}_{CL}= 
-\frac{1}{N} \sum_{i=1}^{N} \log \frac{\exp (\mathcal{S}(v_i,t_l^{'}) / \tau )}{\sum_{l=1}^{L} \exp (\mathcal{S}(v,t_l) / \tau )},
\end{array}
\end{equation} 
where $S$ denotes similarity between extracted visual features $v_i$ and the generated textual conception $t_l$ from embedding set $\{L\}$, $t_l^{'}$ is the most similar textual conception to $v_i$. The model is optimized by large-scale image-text pairs. In this way, the model could learn a strong generalized foreground-background discrimination ability. 

Meanwhile, we also design image-level contrastive loss $\mathcal{L}_{CL\_image}$ for aligning the whole image and corresponding textual description. To optimize the overall pre-training, the overall loss function is described as $\mathcal{L}_{Pre-train}= \mathcal{L}_{CL}+\mathcal{L}_{CL\_image}$. The pre-training is designed to learn feature representations of a large range of object categories, not just limited pre-defined object categories. Through this pre-training, we could introduce redundant knowledge of wide object categories and promote the ability to recognize the foreground object categories from the background.

\subsection{Marine-specific Training}
After the pre-training procedure, the trained model 
by redundant image-text pairs (containing various object categories) could contain generalized foreground-background discrimination ability. Different from existing close-set object detection algorithms, we replace the classifier with textual category prototypes, which could significantly improve generalization to novel categories under the OMOD setting. We use cosine similarity to calculate the matching probability between regional feature $v_i$ and textual conception $t_l^{'}$ in the training:
\begin{equation}
    P(t_l^{'}|v_i)=\frac{\exp{Sim(t_l^{'},v_i)}}{\sum_{l \in \{O\}}\exp{Sim(t_l,v_i)}},
\end{equation}
where $Sim()$ is Cosine similarity function, and $t_l^{'}$ is specific category in $O$. We assign the category label to the image regions within the generated bounding box based on the maximum matching probability. However, directly adopting the pre-trained model for detector initialization and feature extraction for OMOD faces two tricky challenges: 1) the huge conception distance between the in-air object categories presented in CLIP and marine object categories; and 2) the appearance shift from the in-air condition to the underwater/marine conditions. Since the CLIP model is mainly optimized by general-purpose image-text pairs, there are relatively few images containing marine object conceptions. Furthermore, the RPN module also shows a limited ability to extract meaningful foregrounds from marine images. We support these two claims in Sec.~\ref{sec:expriments}. To address these issues, we propose a novel \textbf{MarineDet dataset}, which is specially designed for OMOD. We fine-tune the whole detection network (initialized from the pre-trained model) and the RPN module to the marine domain. To obtain accurate and robust region features, we have also preserved the detection loss function $\mathcal{L}_{Det}$ from Faster R-CNN~\cite{NIPS2015}. Through an \textbf{in-air-to-marine} knowledge transfer, our MarineDet could effectively perform domain-specific object detection. More importantly, we argue that such an in-air-to-marine fine-tuning could achieve better performance than the existing object detection algorithms initialized from ImageNet pre-trained weights. The detector could be better initialized with a strong foreground-background discrimination ability and shared feature extraction ability (\emph{e.g.}, color, texture, shape, and \textit{etc}).

\section{Experiments}

\subsection{MarineDet Dataset}
\begin{figure}[t]
  \begin{center}
    \includegraphics[width=\linewidth]{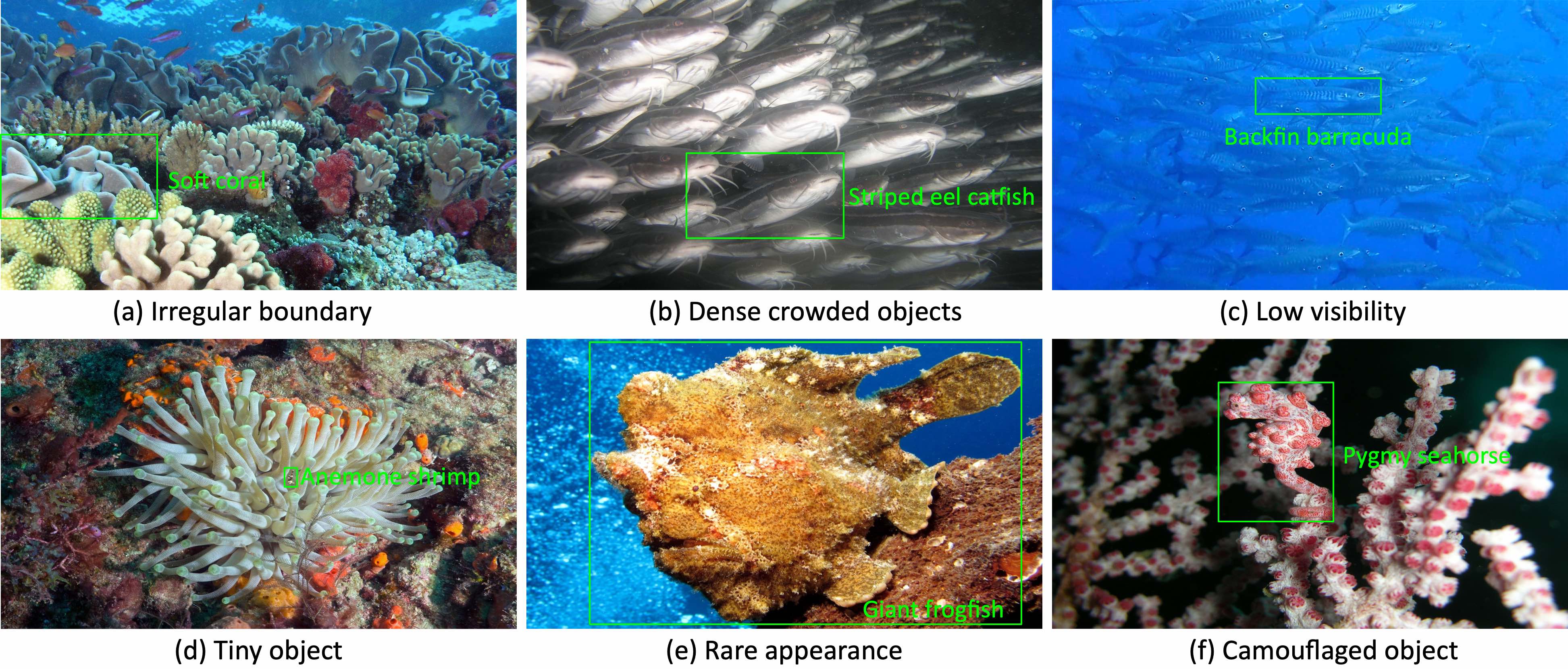}
  \end{center}
  \caption{We illustrate some images from our MarineDet dataset. For images with more than one single object, we only visualize one object instance for better illustration.}
  \label{fig:img_demo}
\end{figure}

Considering there is no comprehensive dataset designed for OMOD, in this work, we propose our \textbf{MarineDet dataset}, which contains 821 diverse object categories. We have also included non-organism categories (\emph{e.g.}, underwater wrecks, surveying devices, sculptures, and \textit{etc}). To include a diverse and comprehensive marine-relative object category list, we ask ChatGPT-3.5/GPT-4 to generate the category candidates and we remove the duplicates. For each category, we manually download relative images from the Internet or crawl the required images from the Google image engine and Flickr website. During the annotation procedure, we manually label these images with dense bounding box annotations. For the images from each category, we \textbf{only} label the objects that belong to the defined object category while ignoring other object categories. In other words, we only label one dominant object category for each image. We provide visualization of some images (with large illumination, visibility, and diversity variations) from our MarineDet dataset in Fig.~\ref{fig:img_demo}. Since our images are crawled from public websites, there are inevitably some noisy images. We remove these noisy images and make sure every category contains at least 10 images. We provide the statistics of our MarineDet dataset and a direct comparison with the existing marine object recognition datasets in Table~\ref{table:statis}. Our MarineDet dataset is the first dataset that contains a large range of marine object conceptions.  

\begin{table*}
    \caption{}
    \vspace{-0.07in}
    \caption*{A direct comparison between our MarineDet dataset with existing marine object recognition datasets.\label{table:statis}}
	\centering
\resizebox{0.9\linewidth}{!}{
\begin{tabular}{c|c|c|c|c|c|c|c}
    \toprule
    Datasets  & Categories & Images & Annotation & No-organism & Diversity & Camouflaged & Task
    \\ \midrule 
     URPC2018 & 4 & 3,701 & BBOX & $\times$ & Medium & $\times$ & Underwater robotics contest\\
     DUO~\cite{liu2021dataset} & 4 & 7,782 & Mask & $\times$ & Medium & $\times$ & Underwater robot picking  \\
     SUIM~\cite{islam2020semantic} & 8 & 1,500 & Mask & $\checkmark$ & Medium & $\times$ & Underwater scene segmentation \\
     MAS3K~\cite{li2020mas3k} & 37 & 3,103 & Mask & $\times$ & High & $\checkmark$ & Marine animal segmentation \\
     Wildfish~\cite{zhuang2018wildfish} & 1,000 & 54,459 & Category & $\times$ & High & $\times$ & Fine-grained fish classification   \\
     \midrule
     \rowcolor[gray]{0.84}MarineDet & 821 & 22,679 & BBOX & $\checkmark$ & High & $\checkmark$ & Open-marine object detection\\
    \bottomrule
    \end{tabular}
}
\end{table*}

\noindent\textbf{Instruction-following attribute generation based on ChatGPT-3.5/GPT-4}. To save human power, we generate scalable attribute annotations for each object category based on ChatGPT-3.5/GPT-4 following designed instructions. We generate the hierarchical biological classification of the marine objects, including ``Kingdom'', ``Phylum'', ``Class'', ``Order'', ``Family'', ``Genus'' and ``Species'' annotations. There are 31 object categories that cannot be defined by the ChatGPT-3.5/GPT-4. We choose ``Class'' as the clustering metric to obtain 26 Classes except 31 undefined object categories. Besides, in our MarineDet dataset, some object categories are given the scientific name (\emph{e.g.}, ``Argopecten irrdaians'') rather than the common name (``Atlantic bay scallop''). Considering this, we generated the common name of these 821 object categories. All generated attribute annotations will be used for data split and evaluation. We will release these annotations with the bounding box annotations for more detailed and fine-grained marine object detection. Since all the attribute annotations are generated by ChatGPT-3.5/GPT-4, there are inevitably mistakes and noises.

\subsection{Implementation Details and Experimental Setups}
\noindent\textbf{Implementation details}. There are mainly two procedures in our framework: \textit{Pre-training} and \textit{Training}. \textbf{Pre-training}. In the pre-training procedure, we adopt a pre-trained and \textbf{frozen} CLIP language encoder~\cite{radford2021learning} as the language encoder and a ResNet-50~\cite{he2016deep} network as the visual backbone. The COCO Caption dataset~\cite{lin2014microsoft} is used for contrastive learning. We adopt an SGD optimizer to optimize the model. The model is first trained with a learning rate of $lr = 5e^{-3}$ for 300K iterations, and then $lr = 5e^{-4}$ for another 300K iterations. The whole training process is with a batch size of 16 on 4 Tesla A100 GPUs, which will take about 1 week. In the \textbf{Training} procedure, we adopt the Faster RCNN~\cite{NIPS2015} as our baseline model and initialize the network backbone with the pre-training model. We replace the classifier weight with textual category embedding. We start the training process by setting the learning rate as $lr = 5e^{-3}$ and then decreasing it to $lr = 5e^{-4}$ and $lr = 5e^{-5}$ when appropriate. We train 60k iterations with a batch size of 8 on 2 Tesla A100 GPUs which takes 16 hours. 

\noindent\textbf{Experimental setup}. We mainly perform experiments under two settings: \textbf{fully supervised} and \textbf{open-vocabulary}. Under the former fully supervised setting, all the 821 object categories are used for training. Under the latter open-vocabulary setting, some object categories are regarded as ``\textit{seen}'' categories and other object categories as ``\textit{unseen}'' categories. Under both settings, four-fifths of the images from each category are used for training and the rest of the images are used for evaluation. We construct \textit{seen}/\textit{unseen} split following three settings: 1) Intra-``\textit{Class}'': all the object categories (613 categories) from 20 ``Classes'' are regarded as \textit{seen} categories and other 177 categories from other 6 ``Classes'' as \textit{unseen} categories. 2) Inter-``\textit{Class}'': we choose one object category from every 4 object categories in each ``Class'' as the \textit{unseen} category and the other 3 object categories as \textit{seen} categories. We omit the ``Class'' that contains less than 4 object categories. In total, there are 572 object categories as \textit{seen} categories and other 183 categories as \textit{unseen} categories. 3) ``\textit{Class}''-level: we adopt the same 20 ``Classes'' as \textit{seen} categories and the other 6 ``Classes'' as \textit{unseen} categories. Compared with Intra-``\textit{Class}'' setting, the ``\textit{Class}''-level setting is performing coarse object recognition while the Intra-``\textit{Class}'' setting is performing fine-grained object recognition.

\begin{figure*}[t]
  \centering
  \includegraphics[width=\linewidth]{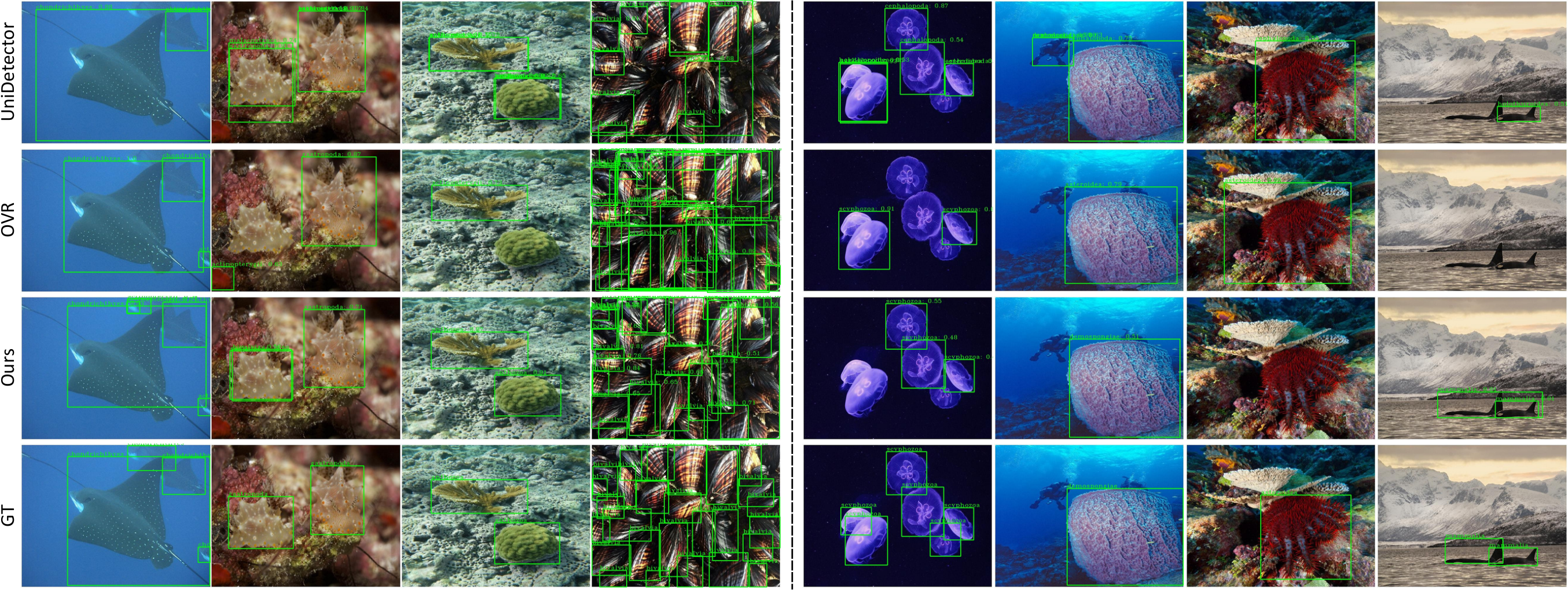}
  \caption{The qualitative comparison between different algorithms. The left part of the dashed line represents the results of \textit{seen} ``Classes'' while the right part shows the results of \textit{unseen} ``Classes''.}
  \label{fig:comp}

\end{figure*}

\subsection{Comparisons with SOTAs}~\label{sec:expriments}

\begin{table*}[!hbt]
    \caption{}
    \vspace{-0.07in}
    \caption*{We report the quantitative result comparison between different object detection algorithms. $-$ indicates that the results cannot be computed.\label{table:26comp}}
	\centering
\scalebox{0.63}{\begin{tabular}{c|c|cccccc|ccccccc}
    \toprule
    \multirow{2}{*}{Method} & \multirow{2}{*}{\begin{tabular}[c]{@{}c@{}}MarineDet\\ Dataset\end{tabular}} & \multicolumn{6}{c|}{Seen (20 ``Classes'')}          & \multicolumn{7}{c}{Unseen (6 ``Classes'')}  \\
    & & mAP$_{50}$ & Actinopterygii & Chondrichthyes & Gastropoda & Malacostraca & Reptilia & mAP$_{50}$ & Cephalopoda & Holothuroidea & Asteroidea & Demospongiae & Mammalia & Scyphozoa \\
    \midrule
    FasterRCNN~\cite{NIPS2015} & \multirow{3}{*}{$\checkmark$} & 41.5 & 58.3 & 47.3 & 49.8 & 50.0 & 64.2  & $-$   &  $-$  &  $-$  &  $-$  & $-$ & $-$  & $-$ \\
    YOLOX~\cite{ge2021yolox} &  &  46.4  & 62.5  &  55.5  & 53.8 &  49.1 &  66.4 & $-$   &  $-$  &  $-$  &  $-$  & $-$ & $-$  & $-$ \\
    GridRCNN~\cite{lu2019grid} &  & 39.1  & 63.2  &  50.3  & 47.7  &  51.3  & 67.0  & $-$   &  $-$  &  $-$  &  $-$  & $-$ & $-$  & $-$ \\
    \midrule
    UniDetector~\cite{wang2023detecting} & \multirow{2}{*}{$\times$}  & 0.9 & 5.6 & 0.5 & 0 & 7.9 & 0.1 & 1.9 & 7.9 & 0 & 0.1 & 0 & 3.2 & 0.1   \\
    GroundingDINO~\cite{liu2023grounding} &  &   0.9    &   8.2   &   2.0   &  1.9    &  2.9    &  0.4    & 9.4 & 27.6  & 3.1 & 1.7 & 1.1  & 19.5 & 3.6  \\
    \midrule
    UniDetector~\cite{wang2023detecting}  & 
    \multirow{3}{*}{$\checkmark$} & 40.9 & 72.0 & 59.1 & 51.3 & 67.8 & 87.8 & 11.6 & \textbf{37.4} & 4.6 & 3.9 & 0.9 & 18.0 & 4.6  \\
    OVR~\cite{zareian2021open} &   & 42.7 & 76.4 & \textbf{62.5} & 55.6 & \textbf{73.9} & 90.1 &  14.5 & 21.0 & 3.3 & \textbf{14.7} & 0.3 & 13.5 & 34.4 \\
    \rowcolor[gray]{0.84}Ours  &  & \textbf{48.3} & \textbf{80.2} & 61.4 & \textbf{59.2} & 73.8 & \textbf{90.1} & \textbf{17.7} & 15.8 & \textbf{4.8} & 3.8 & \textbf{2.0} & \textbf{26.4} & \textbf{53.5}  \\
    \bottomrule
    \end{tabular}}
\end{table*}

We mainly include 3 close-set object detection algorithms (Faster-RCNN~\cite{NIPS2015}, GridRCNN~\cite{lu2019grid} and YOLOX~\cite{ge2021yolox}) and 3 open-vocabulary/open-set object detection algorithms (OVR~\cite{zareian2021open}, GroundingDINO~\cite{liu2023grounding} and UniDetector~\cite{wang2023detecting}) for comparison. We perform experiments under ``\textit{Class}''-level setting, where 20 Classes are regarded as \textit{seen} ``Classes'' and the other 6 Classes as \textit{unseen} ``Classes''. For close-set object detection algorithms, we only report mAP$_{50}$ of 20 seen ``Classes'' and cannot compute the results of other 6 \textit{unseen} ``Classes''. For the three open-vocabulary/open-set object detection algorithms, we compare them under two settings: without and with continuous training on our MarineDet dataset. Under the first setting, considering these object detection generalist models could detect a large range of object conceptions since these models are optimized by large-scale datasets (COCO~\cite{lin2014microsoft}, LVIS~\cite{gupta2019lvis} and ODinW~\cite{li2022elevater}), we directly utilize their released pre-trained models for inference except OVR~\cite{zareian2021open} since the RPN model is not available. Under the second setting, we continuously optimize their pre-trained models on our MarineDet dataset. We do not optimize the GroundingDINO model since the training codes are not available. All the experiments are conducted following the same train/val data split and we report the quantitative results in Table~\ref{table:26comp}. Due to limited space, we only illustrate the AP$_{50}$ of 5 \textit{seen} ``Classes'' and provide detailed results in the appendix. 

We have noticed that the existing generalist object detection algorithms without continuous training only show very limited detection performance as reported in Table~\ref{table:26comp}. The possible reasons could come from 1) the huge conception distance between the in-air object categories and marine object categories; and 2) the appearance shift from the in-air images to underwater images. The pre-trained models only show a limited ability to extract foreground objects. Furthermore, as demonstrated, open-vocabulary/open-set object detection algorithms, when continuously trained on the MarineDet dataset, typically exhibit improved detection performance even on \textit{seen} categories compared to close-set object detection algorithms. We attribute such promoted performance to the optimization through large-scale datasets with redundant supervision during the pre-training phase. Our method could achieve the best object detection performance for both \textit{unseen} and \textit{seen} categories. We provide the qualitative result comparison of OVR, UniDetector, and MarineDet in Fig.~\ref{fig:comp}. UniDetector misrecognized many objects to ``Cephalopoda'' and achieved 37.4 AP$_{50}$ on ``Cephalopoda'' with only low results on other unseen ``Classes''. OVR achieved the highest result on unseen ``Asteroidea''. Our method achieves the highest mAP$_{50}$ on 6 unseen ``Classes'' among all the methods.

\begin{table}[t]
    \caption{}
    \vspace{-0.07in}
    \caption*{We conduct open-vocabulary marine object detection experiments in a fine-grained manner.\label{table:fine}}
	\centering
\scalebox{0.85}{\begin{tabular}{c|c|c}
\toprule
Method        & mAP$_{50}$ (613 \textit{seen} categories) & mAP$_{50}$ (177 \textit{unseen} categories) \\
\midrule
UniDetector~\cite{wang2023detecting} & 21.1  & 0.5 \\
OVR~\cite{zareian2021open}    &  23.6 &  0.9 \\
\rowcolor[gray]{0.84}Ours          &     \textbf{37.7}   &     \textbf{7.3}  \\
\bottomrule
\end{tabular}}
\end{table}

We then conduct experiments in a fine-grained manner. All the defined object categories by ChatGPT-3.5/GPT-4: (613 categories) from the 20 \textit{seen} ``Classes'' are regarded as \textit{seen} categories and the other 177 object categories from the 6 \textit{unseen} ``Classes'' as \textit{unseen} categories. Under this experimental setup, the model is required to perform fine-grained object recognition for the object categories with similar morphologies and appearances. We report the quantitative results of different methods in Table~\ref{table:fine}. Please note all the methods have been trained on our MarineDet dataset. Compared with OVR and UniDetector, our method demonstrates a more robust and stronger ability to detect unseen object categories even under the fine-grained setting. It still remains a challenging task to perform accurate and robust OMOD under a fine-grained setting. 

We have also performed experiments to demonstrate the ability of our MarineDet could effectively detect marine objects even compared with those specialist models that are designed for detecting pre-defined object categories. We compare Faster-RCNN~\cite{NIPS2015}, YOLOX~\cite{ge2021yolox} and GridRCNN~\cite{lu2019grid} on URPC dataset. Following the default train/val split of \footnotemark[1], we train Faster-RCNN, YOLOX, and GridRCNN and report the corresponding experimental results in Table~\ref{table:urpc}. For our MarineDet, we first adopt our trained model under ``\textit{Class}''-level setting for inference. Then we fine-tune the model under the open-vocabulary and fully-supervised settings. As reported in Table~\ref{table:urpc}, the proposed MarineDet could achieve satisfactory detection performance for ``Sea urchin'' even without any fine-tuning on domain-specific data. After the fine-tuning, we achieved a large performance improvement compared with Faster-RCNN, YOLOX, and GridRCNN.

\footnotetext[1]{{\scriptsize{\url{https://challenge.datacastle.cn/v3/cmptDetail.html?id=680}}}}
\begin{table}[!hbt]
    \caption{}
    \vspace{-0.07in}
    \caption*{We report the quantitative object detection results with specialist object detection algorithms. $\dagger$ indicates it belongs to \textit{seen} category under our ``\textit{Class}''-level setting.\label{table:urpc}}
	\centering
\scalebox{0.72}{\begin{tabular}{c|c|cccc|c}
\toprule
Method & URPC & Sea urchin$^{\dagger}$ & Scollop$^{\dagger}$ & Starfish  & Sea cucumber &  mAP$_{50}$  \\
\midrule
FasterRCNN~\cite{NIPS2015} & \multirow{3}{*}{$\checkmark$} & 48.0 & 48.3 & 52.4  & 41.5 & 47.5  \\
YOLOX~\cite{ge2021yolox} &  & 51.5  & 52.6 & 54.7  & 45.8 & 51.1  \\
GridRCNN~\cite{lu2019grid} & & 50.8  & 45.6 & 59.0  & 52.5 & 52.0  \\
\midrule
Ours (inference) &  $\times$    &  31.0  &  0.5  &   1.7  &   0.1   & 8.3      \\
Ours (open-vocabulary) &  $\checkmark$    & \textbf{88.6}  & 76.5 &  2.3   & 0.2  &  41.9 \\
\rowcolor[gray]{0.84}Ours (fully-supervised)  &  $\checkmark$   &    86.4 & \textbf{83.8}  & 45.8  & \textbf{66.6} &  \textbf{70.6} \\
\bottomrule
\end{tabular}}
\end{table}

\begin{table}[!hbt]
    \caption{}
    \vspace{-0.07in}
    \caption*{We report the experimental results under different settings and also the detailed category split.\label{table:ablation}}
	\centering
\scalebox{1}{\begin{tabular}{c|c|cc|cc}
 \toprule
\multirow{2}{*}{Exp} & \multirow{2}{*}{Settings} & \multicolumn{2}{c|}{Categories} & \multicolumn{2}{c}{mAP$_{50}$} \\
& & Seen  & Unseen & Seen & Unseen       \\
\midrule
Exp1 & Intra-``\textit{Class}'' &  613 & 177 &  37.7  &  7.3  \\
Exp2 & Inter-``\textit{Class}''  & 572 &  183  &  38.7  &  23.7  \\
Exp3 & ``\textit{Class}''-level  & 20 &  6    & 48.3  &  17.7 \\
\midrule
Exp4 & Fully supervised  &  821 &  0  & 35.9 & $-$ \\
\bottomrule 
\end{tabular}}
\end{table}

\subsection{Ablation Studies}
We aim to explore the boundary of our MarineDet under different settings. The experimental results are reported in Table~\ref{table:ablation}. Under the fully supervised setting, the detection model is required to perform very fine-grained marine object recognition and thus only achieved 35.9 mAP$_{50}$ on 821 object categories. Comparing the experimental results of Exp1 and Exp2, we could achieve 23.7 mAP$_{50}$ on 183 \textit{unseen} marine object categories while only 7.3 mAP$_{50}$ on 177 \textit{unseen} marine object categories. We attribute this phenomenon to the reason that the model could better utilize the relationship between similar marine object categories that belong to the same ``Class''. How to utilize the relationship between different marine species to promote open-marine object detection performance is a new and valuable research direction. Comparing the experimental results of Exp1 and Exp3, it still requires more effort to perform fine-grained marine object detection under the OMOD setting.
\subsection{Discussions and Limitations}
\noindent\textbf{Potential applications}. OMOD could enable comprehensive monitoring of marine species diversity and population dynamics, aiding in ecological research and conservation efforts. Detecting and tracking invasive species in marine environments, allowing for timely intervention and mitigation strategies to protect native ecosystems. The versatility of OMOD has the potential to revolutionize underwater research, from advancing ecological knowledge to informing conservation policies and sustainable management practices.

\noindent\textbf{Limitations}. Despite our state-of-the-art OMOD performance, our method is not without limitations. Since the language encoder is borrowed from CLIP~\cite{radford2021learning}optimized by the general-purpose image-text pairs, the language encoder is not optimized for marine applications. The generated textual semantic embedding cannot always be effective enough for marine conditions, which will lead to the performance degradation of some categories. We could further design an optimized textual semantic space that can better model semantic information of marine data and improve the overall accuracy and robustness of MarineDet.

\section{Conclusion}
In this work, we present MarineDet, the first work to perform OMOD. We have also proposed the MarineDet dataset with 821 marine-relative object categories, which could support open-marine and fine-grained marine object detection. We have included comprehensive experiments under various experimental settings and ablation studies also provide insight on OMOD. We believe our work and the proposed MarineDet dataset could boost the development of open-marine object detection. We pave the way for future open-marine object recognition research in both academic and industrial communities. 

{
 \bibliographystyle{ieeetr}
\bibliography{icra}
}
\end{document}